\documentclass[letter, 10 pt, conference]{ieeeconf}  

\IEEEoverridecommandlockouts % This command is only needed if 
\usepackage[english]{babel}
\usepackage[utf8]{inputenc}
\usepackage{times} % assumes new font selection scheme installed
\usepackage{cite}

\DeclareMathAlphabet{\pazocal}{OMS}{zplm}{m}{n}

\usepackage{hyperref}
\usepackage[nolist]{acronym}

\usepackage{multirow}

\usepackage{siunitx}
\usepackage{graphicx} % for pdf, bitmapped graphics files
\usepackage{epsfig} % for postscript graphics files
\usepackage{pgfplots}
\usepackage{caption}
\usepackage[font=footnotesize]{subcaption}
\usepackage{pgfplots}

\usepackage{float}

\usepackage[export]{adjustbox}

\usepackage{xcolor}
\definecolor{myYellow}{rgb}{0.93,0.69,0.13}
\definecolor{myPurple}{rgb}{0.49,0.18,0.56}
\definecolor{myGreen}{rgb}{0.26 0.72 0.54}

\usepackage{mathtools}
\usepackage{nicefrac}
\usepackage{amsmath}
\usepackage{amssymb}
\usepackage{bm} % to bold symbols

\DeclareMathOperator*{\minimize}{minimize}
\DeclareMathOperator*{\maximize}{maximize}

\usepackage{etoolbox}
\makeatletter%
\AfterPreamble{%
	\usepackage{hyperref}%
	\let\oldhypertarget\hypertarget%
	\renewcommand{\hypertarget}[2]{%
		\oldhypertarget{#1}{#2}%
		\protected@write\@mainaux{}{%
			\string\expandafter\string\gdef%
			\string\csname\string\detokenize{#1}\string\endcsname{#2}%
		}%
	}%
	\newcommand{\myhyperlink}[1]{%
		\hyperlink{#1}{\csname #1\endcsname}%
	}%
}
\makeatother%

\newcounter{Remark}
\newcounter{Definition}
\newcommand{\displayDefinitions}[2][]{%
	\stepcounter{Definition}%
	\textit{Definition~}\hypertarget{#1}{\theDefinition}\textit{~(#2)}%
}

\newcommand{\refDefinitions}[1][]{%
	Def.~\myhyperlink{#1}%
}

\newcounter{Problem}
\usepackage{algorithm}
\usepackage[noend]{algpseudocode}

\makeatletter
\def\BState{\State\hskip-\ALG@thistlm}
\makeatother

\usepackage{tikz}
\pgfdeclarelayer{foreground}
\pgfsetlayers{background, main, foreground}
\usetikzlibrary{quotes, angles, backgrounds, arrows, automata, shapes, positioning, calc, through, spy, decorations.pathreplacing, decorations.markings, arrows.meta, automata, petri}

\tikzset{
    imglabel/.style={
      rectangle,
      inner sep=2pt,
      text=black,
      minimum height=1em,
      text centered,
      fill=white,
      fill opacity=1.0,
      text opacity=1,
      anchor=south west,
    },
  }
\tikzset{
	state/.style={
		rectangle,
		draw=black, very thick,
		minimum height=1.0em,
		text centered,
	},
}
\tikzset{
  on each segment/.style={
    decorate,
    decoration={
      show path construction,
      moveto code={},
      lineto code={
        \path [#1]
        (\tikzinputsegmentfirst) -- (\tikzinputsegmentlast);
      },
      curveto code={
        \path [#1] (\tikzinputsegmentfirst)
        .. controls
        (\tikzinputsegmentsupporta) and (\tikzinputsegmentsupportb)
        ..
        (\tikzinputsegmentlast);
      },
      closepath code={
        \path [#1]
        (\tikzinputsegmentfirst) -- (\tikzinputsegmentlast);
      },
    },
  },
  mid arrow/.style={postaction={decorate,decoration={
        markings,
        mark=at position .5 with {\arrow[#1]{stealth}}
      }}},
}
\graphicspath{{./figures/}}

\newcommand\copyrighttext{%
    \small \begin{center} \color{red} \textcopyright\,Accepted for presentation to the ``Energy Efficient Aerial Robotic Systems" Workshop at ICRA'23, ExCeL, London, UK. Personal use of this material is permitted. Permission from authors must be obtained for all other uses, in any current or future media, including reprinting/republishing this material for advertising or promotional purposes, creating new collective works, for resale or redistribution to servers or lists, or reuse of any copyrighted component of this work in other works. \end{center}}
\newcommand\copyrightnotice{%
	\begin{tikzpicture}[remember picture,overlay]
	\node[anchor=south,yshift=25.6cm] at (current page.south) 
	{\color{red}\fbox{\parbox{\dimexpr\textwidth-\fboxsep-\fboxrule\relax}{\copyrighttext}}};
	\end{tikzpicture}%
}

\title{\copyrightnotice \LARGE \bf Ergonomic Collaboration between Humans and Robots: An Energy-Aware Signal Temporal Logic Perspective} 

\author{Giuseppe Silano$^{1}$, Amr Afifi$^{2}$, Martin Saska$^{1}$, and Antonio Franchi$^{2,3,4}$   
    \thanks{$^1$Faculty of Electrical Engineering, Department of Cybernetics, Czech Technical University in Prague, 12135 Prague, Czech Republic (emails: {\tt\footnotesize \{giuseppe.silano, martin.saska\}@fel.cvut.cz}).} 
    \thanks{$^2$Robotics and Mechatronics Department, Electrical Engineering,  Mathematics, and Computer Science (EEMCS) Faculty, University of Twente, 7500 AE Enschede, The Netherlands (emails: {\tt\small \{a.n.m.g.afifi, a.franchi\}@utwente.nl}).}
    \thanks{$^3$Department of Computer, Control and Management Engineering, Sapienza University of Rome, 00185 Rome, Italy (email: {\tt\small antonio.franchi@uniroma1.it}).}
    \thanks{$^4$LAAS-CNRS, Universit\'{e} de Toulouse, 31000 Toulouse, France (email: {\tt\small antonio.franchi@laas.fr}).}
    \thanks{This work was partially funded by the EU's H2020 AERIAL-CORE grant no. 871479, by the CTU grant no. SGS23/177/OHK3/3T/13, by the Czech Science Foundation (GAČR) grant no. 23-07517S, and by the OP VVV grant no. CZ.02.1.01/0.0/0.0/16 019/0000765.}
}

\begin{document}

\maketitle
\thispagestyle{empty} % plain to see the number of pages
\pagestyle{empty} % plain to see the number of pages

%%% START SECTION ==========================================================

\begin{acronym}
    \acro{AGM}[AGM]{Arithmetic-Geometric Mean}
    \acro{AR}[AR]{Aerial Robot}
    \acro{CoM}[CoM]{Center of Mass}
    \acro{DoF}[DoF]{Degree of Freedom}
    \acro{GTMR}[GTMR]{Generically Tilted Multi-Rotor}
    \acro{HRI}[HRI]{Human-Robot Interaction}
    \acro{ILP}[ILP]{Integer Linear Programming}
    \acro{LSE}[LSE]{Log-Sum-Exponential}
    \acro{MILP}[MILP]{Mixed-Integer Linear Programming}
    \acro{MRAV}[MRAV]{Multi-Rotor Aerial Vehicle}
    \acro{NLP}[NLP]{Nonlinear Programming}
    \acro{NMPC}[NMPC]{Nonlinear Model Predictive Control}
    \acro{pHRI}[pHRI]{physical Human Robot Interaction}
    \acro{ROS}[ROS]{Robot Operating System}
    \acro{STL}[STL]{Signal Temporal Logic}
    \acro{TL}[TL]{Temporal Logic}
    \acro{UAV}[UAV]{Unmanned Aerial Vehicle}
    \acro{UAM}[UAM]{Unmanned Aerial Manipulator}
    \acro{VRP}[VRP]{Vehicle Routing Problem}
    \acro{wrt}[w.r.t.]{with respect to}
\end{acronym}

%%% END SECTION ============================================================

%%% START SECTION ==========================================================

\begin{abstract}

This paper presents a method for designing energy-aware collaboration tasks between humans and robots, and generating corresponding trajectories to carry out those tasks. The method involves using high-level specifications expressed as~\ac{STL} specifications to automatically synthesize task assignments and trajectories. The focus is on a specific task where a~\ac{MRAV} performs object handovers in a power line setting. The motion planner takes into account constraints such as payload capacity and refilling, while ensuring that the generated trajectories are feasible. The approach also allows users to specify robot behaviors that prioritize human comfort, including ergonomics and user preferences. The method is validated through numerical analyses in MATLAB and realistic Gazebo simulations in a mock-up scenario.

\end{abstract}

%%% END SECTION ============================================================

%%% START SECTION ==========================================================

\section{Introduction}
\label{sec:introduction}

In robotics,~\acfp{MRAV} are popular due to their agility, maneuverability, and versatility with onboard sensors. They have various applications, including contactless or physical interaction with their surroundings~\cite{OlleroTRO2022}.~\acp{MRAV} are advantageous in scenarios such as working environments at heights, wind turbines, large construction sites, and power transmission lines~\cite{SilanoRAL2021}. They can act as robotic co-workers, carrying tools and reducing physical and cognitive load on human operators, but ergonomics and safety must be considered~\cite{Afifi2022ICRA, Corsini2022IROS}. However, the use of~\acp{MRAV} in human-robot interaction is limited compared to ground robots. Object handover is also a well-studied topic.

To enable effective collaboration between~\acp{MRAV} and human workers, advanced task and motion planning techniques are required to address ergonomic and safety concerns while minimizing the physical and cognitive demands on human operators.~\acf{STL}~\cite{maler2004FTMATFTS} can provide a framework to express these complex specifications and generate optimal feasible trajectories. 

Handover involves multiple stages: \textit{approach}, \textit{reach}, and \textit{transfer} phases~\cite{Corsini2022IROS, Afifi2022ICRA}. While some previous studies have examined individual phases, e.g.~\cite{Medina2016Humanoids}, there is limited consideration of safety and ergonomics in such approaches as well as energy efficiency. For aerial robot-human collaboration in high-risk environments, it is crucial to include these considerations. Additionally, prior works~\cite{Sisbot2012TRO, Peternel2017Humanoids} have explored the integration of human comfort and ergonomics in robot planning, but none have considered the context of~\acp{MRAV} as co-workers with humans.

Some studies use sensors on~\acp{MRAV} to improve control and planning, with perception-constrained control being a key consideration. For example,~\cite{Corsini2022IROS} proposes a~\ac{NMPC} formulation that incorporates human ergonomics and comfort while enforcing perception and actuation limits. Other research, such as~\cite{Afifi2022ICRA}, uses dynamic programming to ensure safety when controlling an aerial manipulator during physical interactions with a human operator. However, these approaches only consider scenarios with a single operator and do not address energy consumption. Regarding motion planning for human-robot handovers,~\cite{Kshirsagar2019IROS} presents a controller automatically generated from~\ac{STL} specifications, while~\cite{Webster2019ArXiv} uses probabilistic model-checking to validate a controller for safety and liveness specifications. Neither of these addresses the task assignment and trajectory generation problem to enhance energy-aware human-robot ergonomic collaboration for~\acp{MRAV}.

\begin{figure}[tb]
    \centering
    % left - bottom - right - top
    \adjincludegraphics[width=0.55\columnwidth, trim={{0.22\width} {0.025\height} {0.0\width} {0\height}}, clip]{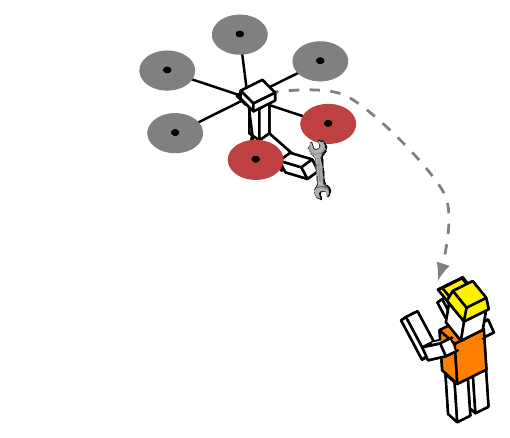}
    \vspace{-0.75em}
    \caption{Illustration of an~\ac{MRAV} approaching a human operator, with gray showing a possible~\ac{STL} optimizer output.}
    \label{fig:scenarioHumanApproach}
\end{figure}

This paper presents an energy-aware motion planner that leverages~\ac{STL} specifications to facilitate human-robot collaboration. To this end, a \textit{nonlinear non-convex max-min} optimization problem is formulated, which is addressed using a hierarchical approach that first solves an~\ac{ILP} problem. The approach is demonstrated in a power line scenario considering the task of an~\ac{MRAV} performing object handovers as depicted in Fig.~\ref{fig:scenarioHumanApproach}, where the mission requirements are expressed as an~\ac{STL} formula. Trajectories consider payload capacity limitations and refilling stations for longer-duration operations. Additionally, a method for computing the initial solution for the optimization problem is proposed. Validation is conducted through numerical simulations in MATLAB, while Gazebo simulations demonstrate the approach's effectiveness in a real-world implementation scenario.

%%% END SECTION ============================================================

%%% START SECTION ==========================================================

\section{Problem Description}
\label{sec:problemDescription}

This paper aims to improve ergonomic human-robot collaboration by designing a trajectory for an~\ac{MRAV} equipped with a manipulation arm to perform object handovers in a power line setting. To meet ergonomic requirements, the drone must approach the operator from the front, either from the left or right, from above or below, and never from behind. Additionally, refilling stations are available for the drone to reload tools. The goal is to complete the mission within a specified maximum time frame while meeting dynamic and capability constraints, as well as avoiding obstacles and minimizing energy consumption. To simplify the scenario, we assume that the handover location is a 3D space for each operator, that the~\ac{MRAV} can carry only one tool at a time, and that an onboard low-level controller, e.g.~\cite{Afifi2022ICRA, Corsini2022IROS}, manages the handover procedure. A map of the environment, including obstacles, is assumed to be known in advance.

%%% END SECTION ============================================================

%%% START SECTION ==========================================================

\section{Preliminaries}
\label{sec:preliminaries}

Let us consider a discrete-time dynamical system of a~\ac{MRAV} $x_{k+1}=f(x_k, u_k)$, where $x_{k+1}$ and $x_k \in \mathcal{X} \subset \mathbb{R}^n$ are the next and current states, respectively, and $u_k \in \mathcal{U} \subset \mathbb{R}^m$ is the control input. Let $f \colon \mathcal{X} \times \mathcal{U} \rightarrow \mathcal{X}$ be differentiable in both arguments. With an initial state $x_0 \in \mathcal{X}_0 \subset \mathbb{R}^n$ and a time vector $\mathbf{t} = (t_0, \dots, t_N)^\top \in \mathbb{R}^{N+1}$, we can define the finite control input sequence $\mathbf{u}=(u_0, \dots, u_{N-1})^\top \in \mathbb{R}^N$ to attain the unique sequence of states $\mathbf{x}=(x_0, \dots, x_N)^\top \in \mathbb{R}^{N+1}$ with sampling period $T_s \in \mathbb{R}^+$ and $N \in \mathbb{N}^+$ samples.

%Let us consider a~\ac{GTMR} model~\cite{Corsini2022IROS} with the world and body frames denoted as $\mathcal{F}_W$ and $\mathcal{F}_B$, respectively, where the origin of $\mathcal{F}_B$ coincides with the~\ac{CoM} of the vehicle indicated as $\mathcal{O}_B$. The position, velocity, and acceleration of $\mathcal{O}_B$ in $\mathcal{F}_W$ are denoted as $\mathbf{p},\mathbf{v},\mathbf{a} \in \mathbb{R}^3$.

Hence, we define the state and control input sequences for the~\ac{MRAV} as $\mathbf{x}=(\mathbf{p}^{(1)}, \mathbf{v}^{(1)}, \mathbf{p}^{(2)}, \mathbf{v}^{(2)}, \mathbf{p}^{(3)}, \mathbf{v}^{(3)})^\top$ and $\mathbf{u}=(\mathbf{a}^{(1)}, \mathbf{a}^{(2)}, \mathbf{a}^{(3)})^\top$, where $\mathbf{p}^{(j)}, \mathbf{v}^{(j)}, \mathbf{a}^{(j)}$ are the position, velocity, and acceleration sequences of the vehicle along the $j$-axis of the world frame $\mathcal{F}_W$, respectively. Finally, let us denote with $p_k^{(j)}, v_k^{(j)}, a_k^{(j)}, t_k$ the $k$-th elements of the sequences $\mathbf{p}^{(j)}, \mathbf{v}^{(j)}, \mathbf{a}^{(j)}$ and vector $\mathbf{t}$, respectively. 

%%% END SECTION ============================================================

%%% START SECTION ==========================================================

\subsection{Signal temporal logic}
\label{sec:signalTemporalLogic}

\displayDefinitions[SignalTemporalLogic]{\acl{STL}}:~\ac{STL} is a concise language for describing real-valued signal temporal behavior~\cite{maler2004FTMATFTS}. Unlike traditional planning algorithms~\cite{LaValleBook}, all mission specifications can be encapsulated into a single formula $\varphi$.~\ac{STL}'s grammar includes temporal operators, such as \textit{until} ($\mathcal{U}$), \textit{always} ($\square$), \textit{eventually} ($\lozenge$), and \textit{next} ($\bigcirc$), as well as logical operators like \textit{conjunction} ($\wedge$), \textit{disjunction} ($\vee$), \textit{implication} ($\implies$), and \textit{negation} ($\neg$). These operators act on atomic propositions, which are simple statements or assertions that are either \textit{true} ($\top$) or \textit{false} ($\bot$). An~\ac{STL} formula $\varphi$ is considered valid if it evaluates to $\top$, and invalid otherwise. More details are available in~\cite{maler2004FTMATFTS, donze2010ICFMATS}. Informally, $\varphi_1 \pazocal{U}_I \varphi_2$ means that $\varphi_2$ must eventually hold within the time interval $I$, while $\varphi_1$ must hold continuously until that point.

%%% ROBUST SIGNAL TEMPORAL LOGIC
\displayDefinitions[RobustSignalTemporalLogic]{\ac{STL} Robustness}: The satisfaction of an~\ac{STL} formula $\varphi$ (\refDefinitions[SignalTemporalLogic]) can be impacted by uncertainties and unexpected events. To ensure a margin of satisfaction, the concept of \textit{robust semantics} for~\ac{STL} formulae has been developed~\cite{maler2004FTMATFTS, donze2010ICFMATS}. This \textit{robustness}, $\rho$, is a quantitative metric that guides the optimization process towards finding the best feasible solution for meeting the statement requirements. It is formally defined using the recursive formulae:
\begin{equation*}
    \begin{array}{rll}
    \rho_{p_i} (\mathbf{x}, t_k) & = & \mu_i (\mathbf{x}, t_k), \\ %\forall p_i \in AP \\
    \rho_{\neg \varphi} (\mathbf{x}, t_k) & = & - \rho_\varphi (\mathbf{x}, t_k), \\
    \rho_{\varphi_1 \wedge \varphi_2} (\mathbf{x}, t_k) & = & \min \left(\rho_{\varphi_1} (\mathbf{x}, t_k), \rho_{\varphi_2} (\mathbf{x}, t_k) \right), \\
    \rho_{\varphi_1 \vee \varphi_2} (\mathbf{x}, t_k) & = & \max \left(\rho_{\varphi_1} (\mathbf{x}, t_k), \rho_{\varphi_2} (\mathbf{x}, t_k) \right), \\
    \rho_{\square_I \varphi} (\mathbf{x}, t_k) & = & \min\limits_{t_k^\prime \in [t_k + I]} \rho_\varphi (\mathbf{x}, t_k^\prime), \\
    \rho_{\lozenge_I \varphi} (\mathbf{x}, t_k) & = & \max\limits_{t_k^\prime \in [t_k + I]} \rho_\varphi (\mathbf{x}, t_k^\prime), \\
    \rho_{\bigcirc_I \varphi} (\mathbf{x}, t_k) & = & \rho_\varphi (\mathbf{x}, t_k^\prime), \text{with} \; t_k^\prime \in [t_k+I], \\
    \rho_{\varphi_1 \mathcal{U}_I \varphi_2} (\mathbf{x}, t_k) & = & \max\limits_{t_k^\prime \in [t_k + I]} \Bigl( \min \left( \rho_{\varphi_2} (\mathbf{x}, t_k^\prime) \right), \\
    & &  \hfill \min\limits_{ t_k^{\prime\prime} \in [t_k, t_k^\prime] } \left( \rho_{\varphi_1} (\mathbf{x}, t_k^{\prime \prime} \right)  \Bigr),
    \end{array}
\end{equation*}
where $t_k + I$ denotes the Minkowski sum of scalar $t_k$ and time interval $I$. The formulae comprise \textit{predicates}, $p_i$, along with their corresponding real-valued function $\mu_i(\mathbf{x}, t_k)$, each of which is evaluated like a logical formula. Namely, $\mathbf{x}$ satisfies the~\ac{STL} formula $\varphi$ at time $t_k$ (in short, denoted as $\mathbf{x}(t_k)\models\varphi$) if $\rho_\varphi(\mathbf{x}, t_k) > 0$, and violates if $\rho_\varphi(\mathbf{x}, t_k) \leq 0$. Each predicate describes part of the mission specifications, and their robustness values indicate how well the specifications are being met. If all predicates are true, then the result is a numerical value that indicates to what degree the specification is being satisfied. Control inputs that maximize robustness are computed over a set of finite state and input sequences, and the optimal sequence $\mathbf{u}^\star$ is considered valid if $\rho_\varphi (\mathbf{x}^\star, t_k)$ is positive.

%%% SMOOTH APPROXIMATION
\displayDefinitions[SmoothApproximation]{Smooth Approximation}: Recent research has proposed smooth approximations $\tilde{\rho}_\varphi(\mathbf{x}, t_k)$ for the non-smooth and non-convex robustness measure $\rho_\varphi(\mathbf{x}, t_k)$, which involves the operators $\min$ and $\max$. These approximations can be optimized efficiently using gradient-based methods. One such smooth approximation is the~\ac{AGM} robustness~\cite{MehdipourACC2019}, which we choose as it is more conservative and computationally efficient than the commonly used~\ac{LSE}~\cite{SilanoRAL2021}. For a full description of the~\ac{AGM} robustness syntax and semantics, see~\cite{MehdipourACC2019}. 

%%% MOTION PLANNER
\displayDefinitions[STL-Planner]{\ac{STL} Motion Planner}: By encoding the mission specifications from Sec.~\ref{sec:problemDescription} as an~\ac{STL} formula $\varphi$ and replacing its robustness $\rho_\varphi(\mathbf{x}, t_k)$ with the smooth approximation $\tilde{\rho}_\varphi(\mathbf{x}, t_k)$ (defined in~\refDefinitions[SmoothApproximation]), the optimization problem for generating energy-aware trajectories for the~\ac{MRAV} can be defined as~\cite{SilanoRAL2021}:
\begin{equation}\label{eq:optimizationProblemMotionPrimitives}
    \resizebox{0.82\hsize}{!}{$%
    \begin{split}
    &\maximize_{ \mathbf{p}^{(j)}, \mathbf{v}^{(j)},\,\mathbf{a}^{(j)}, \bm{\varepsilon}^{(j)} } \;\;
    {\tilde{\rho}_\varphi (\mathbf{p}^{(j)}, \mathbf{v}^{(j)} ) - {\bm{\varepsilon}^{(j)}}^\top \mathbf{Q} \, \bm{\varepsilon}^{(j)}} \\
    &\quad \,\;\, \quad \text{s.t.}~\quad\;\;\; \lvert v^{(j)}_k \rvert \leq 
    \bar{v}^{(j)}, \lvert a^{(j)}_k \vert  \leq \bar{a}^{(j)}, \\
    &\,\;\;\;\;\, \qquad \qquad\;\, \lVert {a_k^{(j)}}^\top a_k^{(j)} \rVert^2 \leq {\bm{\varepsilon}_k^{(j)}}^\top \, \bm{\varepsilon}_k^{(j)}, \, \bm{\varepsilon}^{(j)}_k \geq 0,  \\
    &\,\;\;\;\;\, \qquad \qquad\;\, \mathbf{S}^{(j)}, \forall k=\{0,1, \dots, N-1\}
    \end{split},
    $}%
\end{equation}
where $\bm{\varepsilon} = (\bm{\varepsilon}^{(1)}, \bm{\varepsilon}^{(2)}, \bm{\varepsilon}^{(3)})^\top$ is the sequence of decision variables $\bm{\varepsilon}^{(j)}$ representing the bound on the square norm of the~\ac{MRAV} acceleration along each $j$-axis of $\mathcal{F}_W$. Also, $\bar{v}^{(j)}$ and $\bar{a}^{(j)}$ denote the upper limits of velocity and acceleration, respectively, and $\mathbf{S}^{(j)} ( p_k^{(j)}, v_k^{(j)}, a_k^{(j)} ) = (p_{k+1}^{(j)}, v_{k+1}^{(j)}, a_{k+1}^{(j)})^\top$ are the vehicle motion primitives encoding the splines presented in~\cite{SilanoRAL2021}. The energy minimization pass through the term $\bm{\varepsilon}^\top \mathbf{Q} \, \bm{\varepsilon}$, where $\mathbf{Q} \in \mathbb{R}^{3N \times 3N}$ such that we have $\bm{\varepsilon}^\top \mathbf{Q} \bm{\varepsilon}\geq 0$.

%%% END SECTION ============================================================

%%% START SECTION ==========================================================

\section{Problem Solution}
\label{sec:problemSolution}

% [PREVIOUS_TEXT]: In this section, we apply the~\ac{STL} framework from Sec.~\ref{sec:preliminaries} to formulate the max-min problem presented in Sec.~\ref{sec:problemDescription} as a nonlinear non-convex~\ac{NLP} problem solved through dynamic programming. To solve this type of nonlinear problem in a reasonable amount of time, we generate an initial guess from a simplified~\ac{ILP} formulation, which does not consider obstacle avoidance, safety requirements, vehicle dynamics, ergonomics, energy minimization, or time specifications. This approach facilitates the search for a global solution. The mission requirements, which involve performing object handovers with an~\ac{MRAV} under safety and ergonomic constraints, are translated into the~\ac{STL} formula $\varphi$ considering the mission time $t_N$. The~\ac{STL} formula contains two types of specifications: safety requirements, which ensure that the~\ac{MRAV} stays within a designated area ($\varphi_\mathrm{ws}$), avoids collisions with objects ($\varphi_\mathrm{obs}$), and never approaches the operator from behind ($\varphi_\mathrm{beh}$), and ergonomic-related objectives, which require the~\ac{MRAV} to visit each human operator ($\varphi_\mathrm{han}$), stay with them for a fixed duration $t_\mathrm{han}$, approach them from the front based on their preferences ($\varphi_\mathrm{pr}$), and stop at a refilling station for $t_\mathrm{rs}$ when its onboard supply of tools is depleted. All mission requirements can be expressed using the~\ac{STL} formula.

In this section, we apply the~\ac{STL} framework from Sec.~\ref{sec:preliminaries} to formulate the optimization problem presented in Sec.~\ref{sec:problemDescription} as a nonlinear non-convex max-min problem. To solve this problem, we generate an initial guess using a simplified~\ac{ILP} formulation that does not account for obstacles, safety, vehicle dynamics, ergonomics, energy minimization, or time specifications. This approach simplifies the search for a global solution. We translate the mission requirements, which include performing object handovers with an~\ac{MRAV} under safety and ergonomic constraints, into the~\ac{STL} formula $\varphi$ that considers the mission time $T_N$. The~\ac{STL} formula contains two types of specifications: \textit{safety requirements} that ensure the~\ac{MRAV} stays within a designated area ($\varphi_\mathrm{ws}$), avoids collisions with objects ($\varphi_\mathrm{obs}$), and never approaches the operator from behind ($\varphi_\mathrm{beh}$); and \textit{ergonomic-related objectives} that require the~\ac{MRAV} to visit each human operator ($\varphi_\mathrm{han}$), stay with them for a fixed duration $T_\mathrm{han}$, approach them from the front based on their preferences ($\varphi_\mathrm{pr}$), and stop at a refilling station for $T_\mathrm{rs}$ when its onboard supply of tools is depleted ($\varphi_\mathrm{rs}$). Finally, the~\ac{MRAV} must return to the  refilling station after completing the handover operations ($\varphi_\mathrm{hm}$). All mission requirements can be expressed as:
\begin{equation}\label{eq:stlFormula}
    \resizebox{0.86\hsize}{!}{$%
    \begin{split}
    \varphi =& \, \square_{[0, T_N]} ( \varphi_\mathrm{ws} \wedge \varphi_\mathrm{obs} \wedge \varphi_\mathrm{beh} ) \, \wedge \\
    & \bigwedge_{q=1}^\mathrm{han} \lozenge_{[0,T_N-T_{\mathrm{han}}]} \left( \bigwedge_{d=1}^\mathrm{pr} {^{q,d}}\varphi_{\mathrm{pr}} \wedge  \square_{[0,T_{\mathrm{han}}]} {^q}\varphi_{\mathrm{han}} \right) \, \wedge \\
    & \bigvee_{q=1}^\mathrm{rs} \lozenge_{[0,T_N-T_{\mathrm{rs}}]} \left(c(t) = 0 \implies \mathbf{p}(t)\models {^q}\varphi_{\mathrm{rs}}\right) \, \wedge \\
    & \bigvee_{q=1}^\mathrm{rs} \square_{[1,T_N-1]} \left(\mathbf{p}(t)\models \varphi_{\mathrm{hm}}\Longrightarrow \mathbf{p}(t+1)\models \varphi_{\mathrm{hm}} \right).
    \end{split}
    $}%
\end{equation}
with
\begin{subequations}\label{eq:STLcomponents}
    \begin{align}
    \textstyle{\varphi_\mathrm{ws}} &= \textstyle{\bigwedge_{j=1}^3 \mathbf{p}^{(j)} \in (\underline{p}^{(j)}_\mathrm{ws}, \bar{p}^{(j)}_\mathrm{ws})}, \label{subeq:belongWorkspace} \\
    \textstyle{\varphi_\mathrm{obs}} &= \textstyle{ \bigwedge_{j=1}^3\bigwedge_{q=1}^{\mathrm{obs}} \mathbf{p}^{(j)} \hspace{-0.25em} \not\in ({^q}\underline{p}_{\mathrm{obs}}^{(j)}, {^q}\bar{p}_{\mathrm{obs}}^{(j)})}, \label{subeq:avoidObostacles} \\
    \textstyle{\varphi_\mathrm{beh}} &= \textstyle{ \bigwedge_{j=1}^3}\bigwedge_{q=1}^{\mathrm{beh}} \mathbf{p}^{(j)} \hspace{-0.25em} \not\in ({^q}\underline{p}_{\mathrm{beh}}^{(j)}, {^q}\bar{p}_{\mathrm{beh}}^{(j)}), \label{subeq:behind} \\
    \textstyle{\varphi_\mathrm{hm}} &= \textstyle{\bigwedge_{j=1}^3 \mathbf{p}^{(j)} \hspace{-0.25em} \in (\underline{p}^{(j)}_\mathrm{rs}, \bar{p}^{(j)}_\mathrm{rs})}, \label{subeq:backHome}    
    \end{align}
\end{subequations} 
\vspace{-1.5em}
\begin{subequations}
    \begin{align}
    \textstyle{{^q}\varphi_{\mathrm{han}}} &=  \textstyle{\bigwedge_{j=1}^3 \mathbf{p}^{(j)} \hspace{-0.25em} \in ({^q}\underline{p}^{(j)}_{\mathrm{han}}, {^q}\bar{p}^{(j)}_{\mathrm{han}})}, \label{subeq:handover} \tag{3e} \\
    \textstyle{{^{q}}\varphi_\mathrm{rs}} &= \textstyle{\square_{[0, T_\mathrm{rs}]} \bigwedge_{j=1}^3 \mathbf{p}^{(j)} \hspace{-0.25em} \in ({^q}\underline{p}_\mathrm{rs}^{(j)}, {^q}\bar{p}_\mathrm{rs}^{(j)})}, \label{subeq:refill} \tag{3f} \\
    \textstyle{{^{q,d}}\varphi_{\mathrm{pr}}} &=  \textstyle{\bigwedge_{j=1}^3 \mathbf{p}^{(j)} \hspace{-0.25em} \in ({^{q,d}}\underline{p}^{(j)}_{\mathrm{pr}}, {^{q,d}}\bar{p}^{(j)}_{\mathrm{pr}})}. \label{subeq:preferences} \tag{3g} 
    \end{align}
\end{subequations} \setcounter{equation}{3}   
Equation~\eqref{subeq:belongWorkspace} constrains the~\ac{MRAV}'s position to remain within the workspace, with minimum and maximum values denoted by $\underline{p}^{(j)}_\mathrm{ws}$ and $\bar{p}^{(j)}_\mathrm{ws}$, respectively. Equations~\eqref{subeq:avoidObostacles}, \eqref{subeq:behind}, \eqref{subeq:backHome}, \eqref{subeq:handover}, \eqref{subeq:refill}, and \eqref{subeq:preferences}  provide guidelines for obstacle avoidance, operator safety, mission completion, handover operations, payload capacity, and human operators' preferences, respectively. The payload capacity is represented by $c(t) \in \{0, 1\}$. The vertices of rectangular regions identifying obstacles, areas behind the operators, operators themselves, refilling stations, and human operators' preferences are represented by ${^q}\underline{p}_\mathrm{obs}^{(j)}$, ${^q}\underline{p}_\mathrm{beh}^{(j)}$, $\underline{p}_\mathrm{hm}^{(j)}$, ${^q}\underline{p}_\mathrm{rs}^{(j)}$, ${^{q,d}}\underline{p}^{(j)}_\mathrm{pr}$, ${^q}\bar{p}_\mathrm{obs}^{(j)}$, ${^q}\bar{p}_\mathrm{beh}^{(j)}$, $\bar{p}_\mathrm{hm}^{(j)}$, ${^q}\bar{p}_\mathrm{rs}^{(j)}$, and ${^{q,d}}\bar{p}^{(j)}_\mathrm{pr}$, respectively. % Equation~\eqref{subeq:preferences} accounts for the human operators' preferences for the~\ac{MRAV}'s approach. The lower and upper limits for these regions are denoted as $\underline{p}^{(j)}_\mathrm{pr}$ and $\bar{p}^{(j)}_\mathrm{pr}$, respectively. 

%%% END SECTION ============================================================

%%% START SECTION ==========================================================

\subsection{Initial guess}
\label{sec:initialGuess}

The resulting nonlinear, non-convex max-min problem is solved using dynamic programming, which requires a well-chosen initial guess to avoid local optima~\cite{Bertsekas2012Book}. The strategy for obtaining an appropriate initial guess for the~\ac{STL} motion planner involves simplifying the original problem to an optimization problem with fewer constraints. The resulting~\ac{ILP} problem assigns human operators to the vehicle and provides a navigation sequence for the~\ac{MRAV}. The initial guess considers mission requirements and~\ac{MRAV} payload capacity and refilling operations ($\varphi_\mathrm{hm}$, $\varphi_\mathrm{han}$ and $\varphi_\mathrm{rs}$), but disregards safety and ergonomy requirements ($\varphi_\mathrm{ws}$, $\varphi_\mathrm{obs}$, $\varphi_\mathrm{beh}$, and $\varphi_\mathrm{pr}$), and mission time intervals ($T_N$, $T_\mathrm{han}$ and $T_\mathrm{rs}$).

The graph used to formulate the~\ac{ILP} is defined by the tuple $G = (\mathcal{V}, \mathcal{E}, \mathcal{W}, \mathcal{C})$, where $\mathcal{V}$ is the set of vertices, consisting of human operators ($\mathcal{T}$), refilling stations ($\mathcal{R}$), and the depot ($\mathcal{O}$) where the~\ac{MRAV} is initially located. The number of elements in $\mathcal{T}$, $\mathcal{R}$, and $\mathcal{O}$ are represented by $\tau$, $r$, and $\delta$, respectively. The set of edges and their associated weights are represented by $\mathcal{E}$ and $\mathcal{W}$, respectively, where edge weights are modeled using Euclidean distances. To represent the number of times an edge is selected in the~\ac{ILP} solution, an integer variable $z_{ij} \in \mathbb{Z}_{\geq 0}$ is defined for each edge $e_{ij} \in \mathcal{E}$. The variable $z_{ij}$ is limited to the set $\{0,1\}$ if $\{i,j\} \in \{ \mathcal{T},\mathcal{O} \}$ and $\{0,1,2\}$ if $i \in \mathcal{R}$ and $j \in \mathcal{T}$, which ensures that an edge between two human operators is never traversed twice and that the depot is only used as a starting point. The~\ac{ILP} problem is then formulated as:
\begin{subequations}\label{eq:ILP}
    \begin{align}
        % Cost function
        &\minimize_{z_{ij}}
        { \sum\limits_{ \{i,j\} \in \mathcal{V}, \, i \neq j} \hspace{-1em} w_{ij} \, z_{ij} } \label{subeq:objectiveFunction} \tag{4a} \\
        %
        % Constraints (subject to part)
        &\quad \;\; \text{s.t.} \;\, \hspace{-0.15cm} \sum\limits_{i \in \mathcal{V}, \, i \neq j} z_{ij} = 2, \; \forall j \in \mathcal{T}, \label{subeq:visitedOnce} \tag{4b} \\ % Codify that each node must be visited once
        %
        %&\qquad \, \;\,\, \sum\limits_{ i \in \mathcal{V}, \, i \neq j} \quad \hspace{-1.05em} z_{ij} = 2y_{j}, \; \forall j \in \mathcal{T}, \label{subeq:visitedOneUAV} \\ % Codify each node must be visited only by one UAV
        %
        &\qquad \;\,\;\;\;\; \sum\limits_{ i \in \mathcal{T} } \quad \hspace{-0.285em} z_{0i} = 1, \label{subeq:depotVisitedOnce} \tag{4c} \\ % The depot must be visited once by UAV (start mission)
        &\qquad \;\; \sum\limits_{\substack{ i \in \mathcal{T}, \, j \not\in \mathcal{T}}} \quad \hspace{-1.15em} z_{ij} \geq 2 h\hspace{-0.2em}\left(\mathcal{T}\right). \label{subeq:capacityAndSubtours} \tag{4d} % Capacity constraint (generalized subtour elimination constraint) %\; \forall \mathcal{S} \subset \mathcal{T},  
     \end{align}
\end{subequations}
In the formulated~\ac{ILP} problem, the objective function~\eqref{subeq:objectiveFunction} minimizes the distance traversed by the~\ac{MRAV}. Constraints~\eqref{subeq:visitedOnce}, ~\eqref{subeq:depotVisitedOnce} and~\eqref{subeq:capacityAndSubtours} ensure that each human operator is visited once, the~\ac{MRAV} begins at the depot and does not return, tours do not exceed payload capacity or are not connected to a refilling station using $h\hspace{-0.2em}\left(\mathcal{T}\right)$~\cite{Miller1960JACM}, respectively. The motion primitives for the~\ac{MRAV} are obtained from the optimal assignment, which is used to generate a dynamically feasible trajectory. The trajectory includes time intervals for handover and refilling ($T_\mathrm{han}$ and $T_\mathrm{rs}$), with fixed rest-to-rest motion between operators and maximum values for velocity and acceleration ($\bar{v}^{(j)}$ and $\bar{a}^{(j)}$). Further details on the motion primitives are provided in~\cite{SilanoRAL2021}.

%%% END SECTION ============================================================

%%% START SECTION ==========================================================

\section{Simulation Results}
\label{sec:simulationResults}

Numerical simulations in MATLAB were used to validate the planning approach, without including vehicle dynamics and trajectory tracking controller. Feasibility was verified in Gazebo with software-in-the-loop simulations~\cite{Baca2020mrs}. The~\ac{ILP} problem was formulated using the CVX framework, and the~\ac{STL} motion planner used the CasADi library with IPOPT as the solver. Simulations were run on an i7-8565U processor with $32$GB of RAM on Ubuntu 20.04. Illustrative videos with the simulations are available at~\url{http://mrs.felk.cvut.cz/stl-ergonomy-energy-aware}.

The object handover scenario outlined in Sec.~\ref{sec:problemDescription} was used to evaluate the proposed planning strategy. The simulation scenario consisted of a mock-up environment, with two human operators, one refilling station, and a single~\ac{MRAV}. Parameters and corresponding values used to run the optimization problem are listed in Table~\ref{tab:tableParamters}. The heading angle of the~\ac{MRAV} was adjusted by aligning the vehicle with the direction of movement when moving towards the human operator. Once the~\ac{MRAV} reaches the operator, it is assumed that an onboard low-level controller, e.g.~\cite{Afifi2022ICRA, Corsini2022IROS}, handles the handover operation, thus adjusting the heading angle accordingly.  
The rectangular regions in which the~\ac{MRAV} was allowed to approach the operator were established taking into consideration the operators' heading, $\psi_\mathrm{ho1}$ and $\psi_\mathrm{ho2}$, as well as their preferred direction of approach ($\varphi_\mathrm{pr}$).

\begin{table}[tb]
    \centering
	\begin{adjustbox}{max width=1\columnwidth}
	\begin{tabular}{|c|c|c|c|c|c|}
        \hline
        \textbf{Parameter} & \textbf{Symbol} & \textbf{Value} & \textbf{Parameter} & \textbf{Symbol} & \textbf{Value} \\
        \hline
	Max. vel. and acc. & $\{\bar{v}^{(j)}, \bar{a}^{(j)}\}$ & $\SI{1.1} {[\meter\per\square\second]}$ & Mission time & $T_N$ & $\SI{23}{[\second]}$ \\
        Handover time & $T_\mathrm{han}$ & $\SI{3}{[\second]}$ & Refilling time & $T_\mathrm{rs}$ & $\SI{3}{[\second]}$ \\
        Sampling period & $T_s$ & $\SI{0.05}{[\second]}$ & Number of samples & $N$ & $\SI{460}{[-]}$ \\
        Heading operator HO1 & $\psi_{\mathrm{ho1}}$ & $\si{\pi} \si{[\radian]}$ & Heading operator HO2 & $\psi_{\mathrm{ho2}}$ & $\si{0} \si{[\radian]}$ \\
	\hline
	\end{tabular}
	\end{adjustbox}
        \vspace{-0.5em}
	\caption{Parameter values for the optimization problem.}
	\label{tab:tableParamters}
\end{table}

\begin{figure}[tb]
    \centering	
    %\vspace{-0.50em}
\hspace{-1.525cm}
\begin{subfigure}{0.45\columnwidth}
    \centering
    \scalebox{0.52}{
    %%%%%%%%%%%%%%%%%%%%% MRAV NO ENERGY - FRONT, LEFT, RIGHT TRAJECTORIES %%%%%%%%%%%%%%%%%%%%%
    \begin{tikzpicture}
    \begin{axis}[%
    width=2.8119in,%
    height=1.8183in,%
    at={(0.758in,0.481in)},%
    scale only axis,%
    xmin=0,%
    xmax=23,%
    ymax=1.0,%
    ymin=0.0,%
    xmajorgrids,%
    ymajorgrids,%
    ylabel style={yshift=-0.555cm, xshift=-0.15cm}, %shifting the y line text
    xlabel={Time [\si{\second}]},%
    extra x ticks={33.50,38.50,50.50,55.50,67.00,79.00,115.00,120.00,131.00,143.00},%
    extra x tick labels={ , ,  , },%
    ylabel={[\si{J}]},%
    ytick={0,0.2,0.4,0.6,0.8,1},%
    %xtick={0,2,...,12},%
    %xticklabels={ , , },%
    %xtick={0,50,100,150},%
    axis background/.style={fill=white},%
    legend style={at={(0.5,0.15)},anchor=north,legend cell align=left,draw=none,legend 
        columns=-1,align=left,draw=white!15!black}
    ]
    \addplot [color=blue, dotted, line width=1.25pt] file{matlabPlots/experimental_data/square_norm_UAV1_leftRight_energy.txt};%
    \addplot [color=red, dashed, line width=1.25pt] file{matlabPlots/experimental_data/square_norm_UAV1_topDown_energy.txt};%
    \addplot [color=green, dash dot, line width=1.25pt] file{matlabPlots/experimental_data/square_norm_UAV1_LP_energy.txt};%
    %
    %\legend{$\mathrm{drone1}$, $\mathrm{drone2}$};%
    \end{axis}
    \end{tikzpicture}
    }
\end{subfigure}
\hspace{0.35cm}
\begin{subfigure}{0.30\columnwidth}
    \centering
    \scalebox{0.52}{
    %%%%%%%%%%%%%%%%%%%%% MRAV ACCOUNTING FOR ENERGY - FRONT, LEFT, RIGHT TRAJECTORIES %%%%%%%%%%%%%%%%%%%%%
    \begin{tikzpicture}
    \begin{axis}[%
    width=2.8119in,%
    height=1.8183in,%
    at={(0.758in,0.481in)},%
    scale only axis,%
    xmin=0,%
    xmax=23,%
    ymax=1.0,%
    ymin=0.0,%
    xmajorgrids,%
    ymajorgrids,%
    ylabel style={yshift=-0.555cm, xshift=-0.15cm}, %shifting the y line text
    xlabel={Time [\si{\second}]},%
    extra x ticks={33.50,38.50,50.50,55.50,67.00,79.00,115.00,120.00,131.00,143.00},%
    extra x tick labels={ , ,  , },%
    ylabel={[\si{J}]},%
    ytick={0,0.2,0.4,0.6,0.8,1},%
    %xtick={0,2,...,12},%
    %xticklabels={ , , },%
    %xtick={0,50,100,150},%
    axis background/.style={fill=white},%
    legend style={at={(0.5,0.15)},anchor=north,legend cell align=left,draw=none,legend 
        columns=-1,align=left,draw=white!15!black}
    ]
    \addplot [color=blue, dotted, line width=1.25pt] file{matlabPlots/experimental_data/square_norm_UAV1_leftRight.txt};%
    \addplot [color=red, dashed, line width=1.25pt] file{matlabPlots/experimental_data/square_norm_UAV1_topDown.txt};%
    \addplot [color=green, dash dot, line width=1.25pt] file{matlabPlots/experimental_data/square_norm_UAV1_LP.txt};%
    %
    %\legend{$\mathrm{drone1}$, $\mathrm{drone2}$};%
    \end{axis}
    \end{tikzpicture}
    }
\end{subfigure}
    \vspace{-2.25mm}
    \caption{Normalized energy consumption profiles considering different operators’ preferred approach directions, including left and right (blue), front (green), and top to bottom (red). From left to right: the data with and without considering the energy term in the~\ac{STL} motion planner.}
    \label{fig:comparisionEnergyScenarios}
\end{figure}
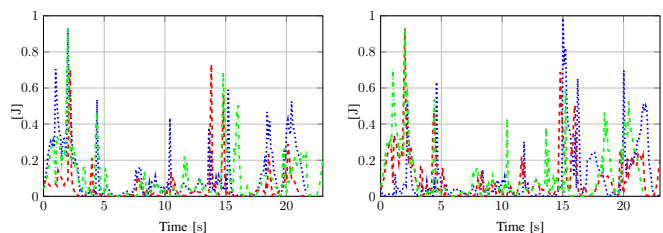

Figure~\ref{fig:comparisionEnergyScenarios} presents a comparison of energy profiles obtained by considering the preferred approach directions of the operators, namely front, right and left, and top to bottom, both with and without the energy term. The energy term is given by $\bm{\varepsilon}_k^\top \mathbf{Q} \bm{\varepsilon}_k \geq 0$ and $\lVert {a_k^{(j)}}^\top a_k^{(j)} \rVert^2 \leq {\bm{\varepsilon}^{(j)}}^\top \, \bm{\varepsilon}^{(j)}, \bm{\varepsilon}^{(j)} \geq 0$, as formulated in the problem statement~\eqref{eq:optimizationProblemMotionPrimitives}. The results demonstrate that the inclusion of the energy term leads to a reduction of energy consumption by approximately 10\%.

%%% END SECTION ============================================================

%%% START SECTION ==========================================================

\section{Conclusions}
\label{sec:conclusions}

This paper presented a motion planning framework to improve energy-aware human-robot collaboration for an~\ac{MRAV} with payload limitations and dynamic constraints. The proposed approach uses~\ac{STL} specifications to generate safe and ergonomic trajectories while meeting mission time requirements. An~\ac{ILP} method is introduced to handle the nonlinear non-convex optimization problem. Numerical in MATLAB and realistic simulations in Gazebo confirm the effectiveness of the proposed approach. Future work includes incorporating human operator fatigue and exploring other types of temporal logic languages to adapt the framework for dynamic environments.

\vspace{-0.5em}

%%% END SECTION ============================================================

%%% START SECTION ==========================================================

\bibliographystyle{IEEEtran}
\bibliography{bib_short}

\end{document}